\documentclass[letterpaper, 10 pt, conference]{ieeeconf}

\IEEEoverridecommandlockouts
\usepackage{algorithmic}
\usepackage{graphicx}
\usepackage{textcomp}
\usepackage{xcolor}
\usepackage{titlesec}
\usepackage{multirow}
\usepackage{subfigure}
\usepackage{amsmath,amssymb,amsfonts}
\usepackage{marvosym}
\usepackage{threeparttable}
\usepackage{pifont}
\usepackage{cite}

\makeatletter
\let\NAT@parse\undefined
\makeatother
\usepackage{hyperref}

\begin{document}

\title{\LARGE \bf
Robotic Sim-to-Real Transfer for Long-Horizon Pick-and-Place Tasks in the Robotic Sim2Real Competition\thanks{\textit{This paper has been accepted for presentation at ICRA 2025. The final version will be available in IEEE Xplore.}}
}

\author{Ming Yang$^{1,2*}$, Hongyu Cao$^{3*}$, Lixuan Zhao$^3$, Chenrui Zhang$^3$, Yaran Chen\textsuperscript{4\Letter}
    \thanks{$^{1}$The State Key Laboratory of Multimodal Artificial Intelligence Systems, Institute of Automation, Chinese Academy of Sciences, Beijing 100190, China, yangming2023@ia.ac.cn}
    \thanks{$^{2}$School of Artificial Intelligence, University of Chinese Academy of Sciences, Beijing, China.}
    \thanks{$^{3}$School of Electrical and Information Engineering, Tianjin University, Tianjin, China.}
    \thanks{$^{4}$Xi'an Jiaotong-Liverpool University, Suzhou, China.}
    \thanks{$*$Equal contribution.}
    \thanks{\textsuperscript{\Letter}Corresponding to yaran.chen@xjtlu.edu.cn}
}

\maketitle
\thispagestyle{empty}
\pagestyle{empty}
\begin{abstract}
This paper presents a fully autonomous robotic system that performs sim-to-real transfer in complex long-horizon tasks involving navigation, recognition, grasping, and stacking in an environment with multiple obstacles.

The key feature of the system is the ability to overcome typical sensing and actuation discrepancies during sim-to-real transfer and to achieve consistent performance without any algorithmic modifications. To accomplish this, a lightweight noise-resistant visual perception system and a nonlinearity-robust servo system are adopted.

We conduct a series of tests in both simulated and real-world environments. The visual perception system achieves the speed of 11 ms per frame due to its lightweight nature, and the servo system achieves sub-centimeter accuracy with the proposed controller. Both exhibit high consistency during sim-to-real transfer. Benefiting from these, our robotic system took first place in the mineral searching task of the Robotic Sim2Real Challenge hosted at ICRA 2024.
\end{abstract}

\section{Introduction}

\begin{figure}[t]
\centerline{\includegraphics[width=\columnwidth]{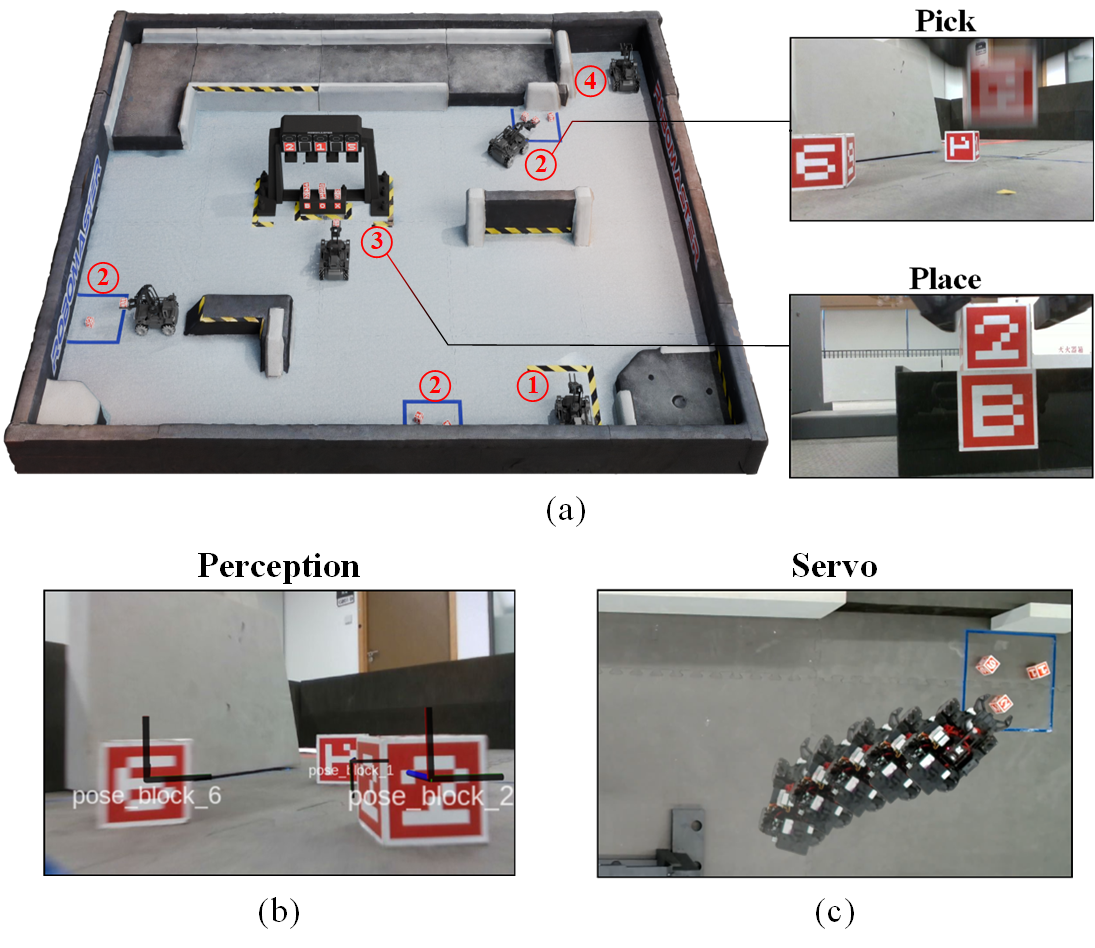}}
\caption{The robot begins at a predefined starting position (see \ding{172} in (a)) and proceeds to search for and grasp "minerals"—cubes marked with unique digits—randomly scattered across three distinct zones (\ding{173}). It then stacks the minerals as high as possible on a platform beneath the exchange station, which is shaped like a door frame with three markers on top indicating the stack order (\ding{174}). The task concludes with the robot stopping at a designated parking point (\ding{175}). Successful picking and placing rely on high-precision visual perception (b) and accurate servo control (c).}
\label{fig:key-feature}
\end{figure}

The pick-and-place task, which integrates key functionalities such as perception and servo control, is a fundamental skill of embodied intelligence. It is critical for a wide range of applications, from autonomous sorting in a warehouse to item organizing by household robots. However, algorithms developed in simulators often face significant discrepancies when transferred to real-world robotic systems, leading to severe performance degradation or even total failure. This challenge is commonly referred to as the sim-to-real gap.

Traditional approaches to addressing the sim-to-real gap typically involve applying reinforcement learning algorithms in simulators \cite{survey-drl-sim2real}, followed by transferring them to real systems with the aid of domain randomization \cite{ex1-dynamics-DR, ex2-dynamics-DR} or transfer learning \cite{ex1-DA, ex2-DA, ex3-DA} in real-world settings. However, the noise introduced by domain randomization is often insufficient for generalization to complex environments, and transfer learning in the real-world often requires extensive interaction data \cite{van2019sim}, which is difficult to obtain.

Therefore, the main purpose of this paper is to demonstrate that \textbf{it is both possible and feasible to achieve consistent performance in long-horizon pick-and-place tasks across simulated and real-world environments without any modification to the algorithm.}

To make this possible, our robotic system consists of two components, with the aim of addressing sensing and actuation discrepancies, respectively. The first is a visual perception system that operates on a detection-classification-localization pipeline with proposed Sequential Motion-blur Mitigation Strategy (SMMS) to minimize the impact of the sensing discrepancy throughout the pipeline. The second is a feedback-linearized servo system with proposed Design Function (DF), which addresses actuation discrepancies of both inappropriate grasp poses (illustrated in Fig. \ref{fig:real-sim grasp}) and unmodeled nonlinearities.

These algorithms were developed as part of the ZeroBug Team system that took first place in the mineral searching task\footnote{AKA, RoboMaster Univerisity Sim2real Challenge, one of the two tracks of RSC.} in the Robotic Sim2Real Challenge (RSC) \cite{RSC2024}. Fig. \ref{fig:key-feature} briefly illustrates the environment and task setup. All code can be accessed via our GitHub repository at \url{https://github.com/Bob-Eric/rmus2024_solution_ZeroBug}, and the simulator is available from the competition committee at \url{https://github.com/AIR-DISCOVER/ICRA2024-Sim2Real-RM}.

The main contributions of this paper are summarized as follows.
\begin{enumerate}
    \item \textbf{High-Consistency Visual Perception System}: We design Sequential Motion-Blur Mitigation Strategy for our visual perception system to handle typical sensing errors, delivering consistent performance in both simulated and real-world environments.
    \item \textbf{Nonlinearity-Robust Servo System}: We introduce Design Function to our feedback-linearized servo system to mitigate actuation discrepancies, demonstrating strong robustness to nonlinearities.
    \item \textbf{Modular System Architecture}: The robotic system features a modular architecture, offering a flexible and adaptable platform for researchers.
\end{enumerate}

\section{Related Works}

In recent years, a variety of simulators have been introduced for embodied intelligence \cite{pyrep,ai2thor,minos,habitat,sapien,house3d}. In these simulators, sim-to-real gap can be broadly categorized into sensing discrepancies (variations in color and texture, non-ideal imaging, etc.) and actuation discrepancies (modeling error, nonlinearities, condition-based simulation of pick-up, etc.) \cite{survey-drl-sim2real}.

\begin{figure}[h]
    \centering
    \subfigure[]{        
        \includegraphics[width=0.22\columnwidth]{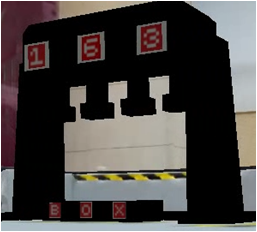}
        \includegraphics[width=0.227\columnwidth]{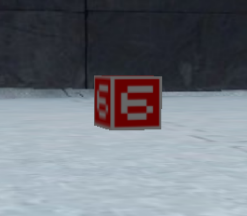}
    }
    \hfill
    \subfigure[]{
        \includegraphics[width=0.22\columnwidth]{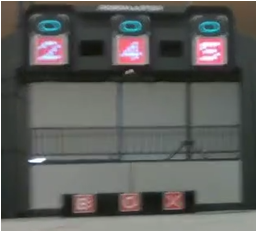}
        \includegraphics[width=0.226\columnwidth]{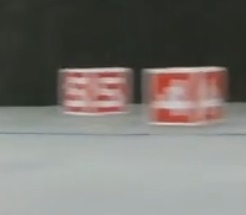}
    }
    \caption{Images of markers captured in the simulator (a) and real-world (b) while robot rotating. Motion blur is difficult to simulate and poses challenges to robotic perception.}
    \label{fig:sensing discrepancies}
\end{figure}

For sensing discrepancies, since Tobin et al. \cite{domain-randomization} first introduce domain randomization, it is widely used in simple and stationary environments \cite{ex1-DR, ex2-DR, ex3-DR}. However, for mobile manipulators with on-board cameras, motion blur, which domain randomization fall short in addressing, may be one of the most common challenges (illustrated in Fig. \ref{fig:sensing discrepancies}). In section \ref{visual perception system}, we propose SMMS to minimize the impact of motion blur.

For actuation discrepancies, popular approaches include reinforcement learning with fine-tuning \cite{ex1-finetune, ex2-finetune, ex3-finetune}, dynamics randomization \cite{ex1-dynamics-DR, ex2-dynamics-DR} and domain adaptation \cite{ex1-DA, ex2-DA, ex3-DA}, all of which require extensive interactions. In section \ref{servo system}, we introduce DF to adapt to reality with minimal interactions.

The 2024 RSC simulator is built with Habitat \cite{habitat}. However, given that the physics and rendering engines used in mainstream simulators are similar, we believe proposed methods for overcoming sensing and actuation discrepancies can be broadly applicable to other simulators.

\section{Challenge I: Perceiving with Sensing Discrepancies} \label{visual perception system}
\begin{figure}[htbp]
\centerline{\includegraphics[width=\columnwidth]{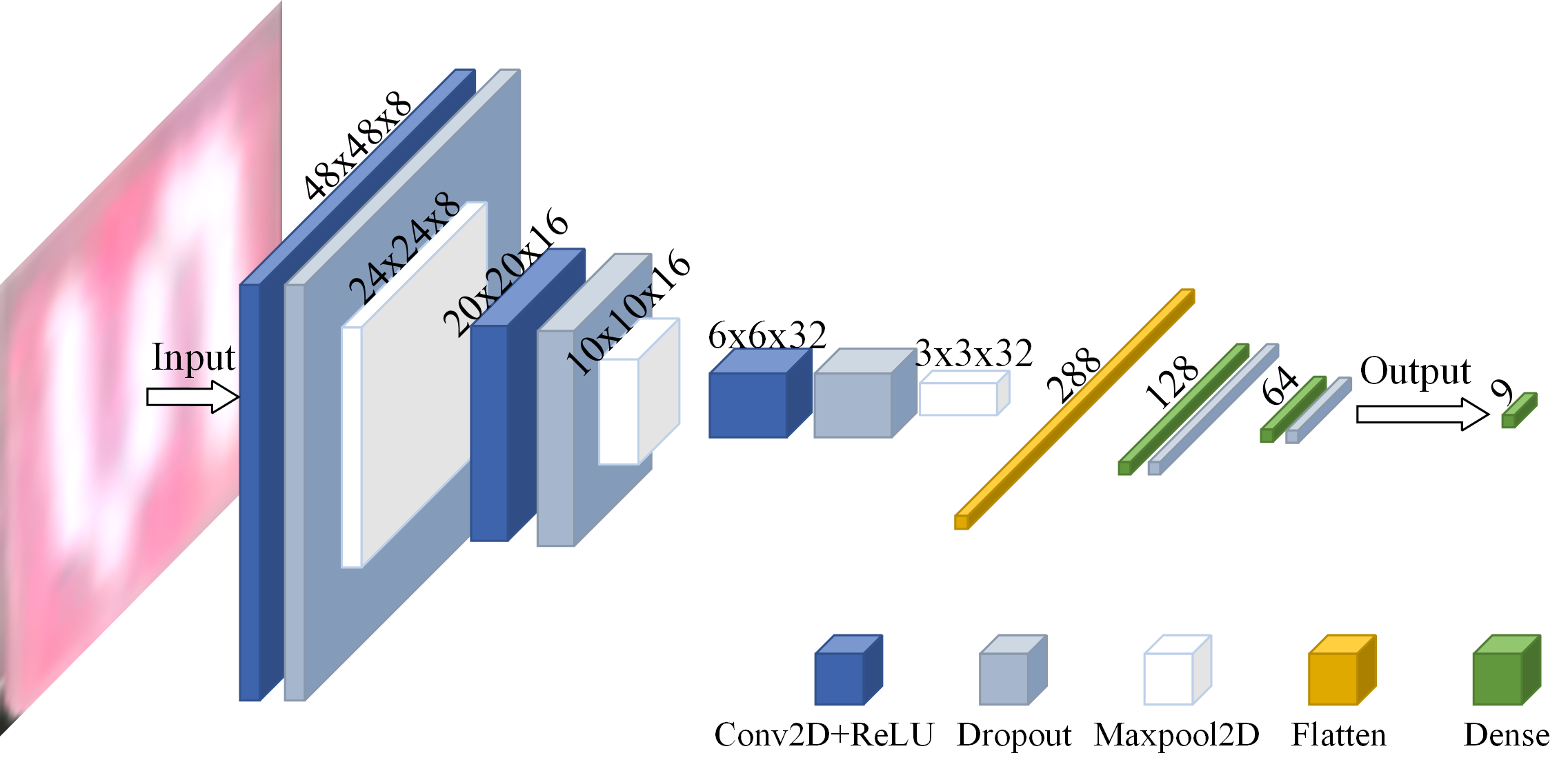}}
\caption{The CNN classifier comprises three convolution-dropout-pooling modules, followed by three fully connected layers. Its lightweight architecture (62.1k parameters in total) enables real-time inference.}
\label{fig:CNN-framework}
\end{figure}

The first step in our system is to achieve consistently precise and real-time pose estimation. Learning-based algorithms \cite{yolov8, faster-rcnn, DL-pose-estimation}, while offering high accuracy and strong robustness in detecting and localizing fiducial markers, often suffer from high computational overhead and extended run-times. In contrast, traditional computer vision techniques (ArUco \cite{aruco, arucoE}, AprilTag \cite{apriltag, apriltag2}) are inherently more efficient but struggle with robustness, particularly in handling noise or image defects. To balance accuracy and efficiency, we replace the template-based classification in the ArUco detector with a CNN-based classifier (illustrated in Fig. \ref{fig:CNN-framework}), drawing on a network architecture similar to LeNet-5 \cite{lenet}.

\begin{figure*}[htbp]
\centerline{\includegraphics[width=\textwidth]{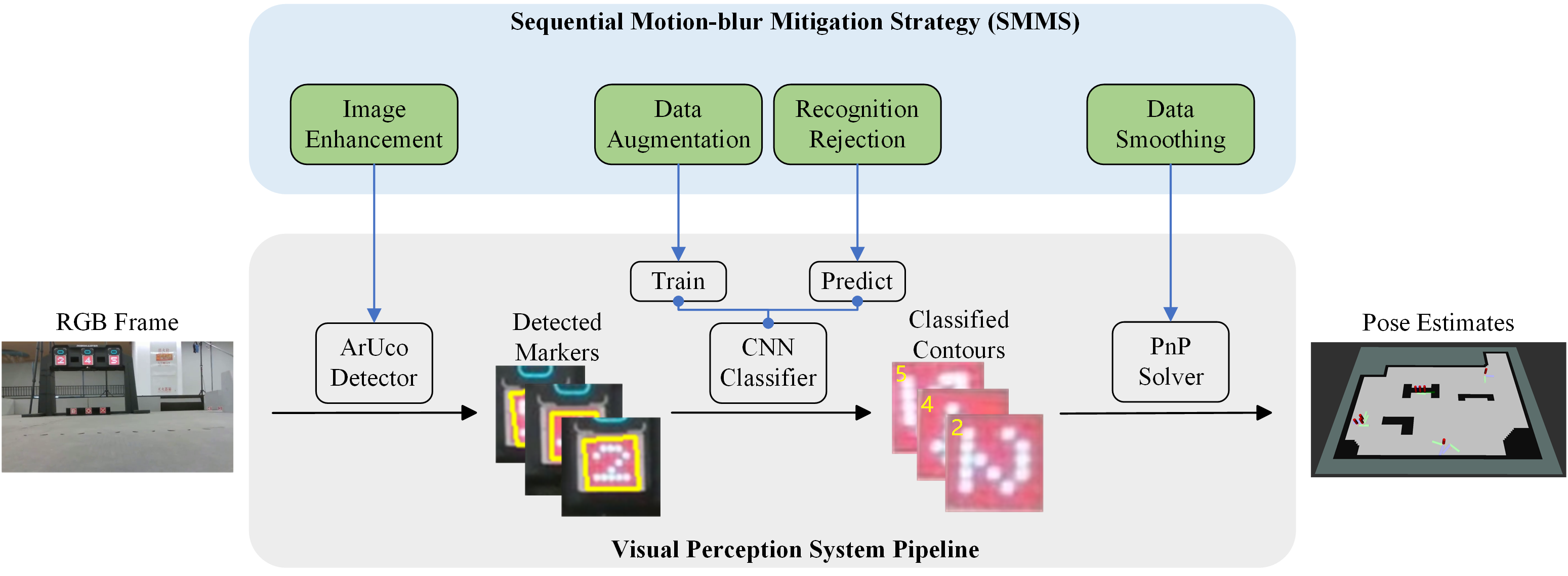}}
\caption{Pipeline of our visual perception system with SMMS. We first use ArUco for extracting contours of markers. Then we rectify each contour with perspective transformation and classify it with CNN classifier, followed by a Perspective-n-Points (PnP) solver to estimate its pose. Throughout the pipeline, SMMS first enhances the image contrast before feeding the feature map into the detector. It then augments training dataset and rejects unreliable classifications. Subsequently, it applies extrapolation and filtering for final pose estimates.}
\label{fig:perception-pipeline}
\end{figure*}

Additionally, motion blur, as the most common sensing discrepancy during sim-to-real transfer, poses great challenges to perceptual performance. It affects pose estimation of each mineral through the following path: motion blur leads to false recognitions and deformed contours, both of which cause imperfect images of markers after perspective transformation, and thereby misclassifications. Subsequently, deformed or misclassified contours introduce noise or abrupt change to pose estimates.
To address this, we propose Sequential Motion-blur Mitigation Strategy (SMMS), illustrated in Fig. \ref{fig:perception-pipeline}.

\textbf{Image Enhancement.} Vanilla ArUco detector \cite{aruco} suffers great loss of recall rate due to motion blur, for which we boost image contrast to compensate. Specifically, we first split RGB channels of the given image and boost the red channel ($R$) with the difference of $R$ and $B$. Then we subtract $G$ from the boosted $R$ to get the feature map (AKA, the input of ArUco detector):
\begin{equation}
    feature = R + k\cdot max(R-B,0) - G
\end{equation}
where $k\in(0,+\infty)$ is an enhancement factor.
Compared with weighted sum form adopted by vanilla ArUco detector ($feature=0.299\cdot R+0.587\cdot G+0.114\cdot B$), it sharpens the boundary of red mineral markers, thereby raising the recall rate from 50.5\% to 70.0\% in real-world (see Table \ref{tab:perception-perf}).

\textbf{Data Augmentation.} We first collect 5 to 9 samples for each class, and then augment the training set with random rotation and color jitter to adapt to reality. Specifically, we first apply random rotations between $\pm$10 degrees to adapt to deformation and skew of transformed images. Then we apply random perturbations to the brightness, contrast, saturation, and hue of each training image.

\textbf{Recognition Rejection.} Incorrect contours produced by the ArUco detector are assigned labels based on the highest score of nine categories, leading to erroneous pose estimation of the corresponding minerals. To address this, we analyze the distribution of unscaled scores of incorrect contours and establish a threshold-based rejection rule. Briefly, it eliminates false recognitions with minimal loss of detector's recall rate.

\textbf{Data Smoothing.} To further mitigate the impact of motion blur, we apply extrapolation for undetected minerals' position and low-pass filter for the detected ones, which compensate for loss of recall rate and precision, respectively. Concretely, with history estimation, we hold poses of undetected minerals in current video frame, and filter poses of detected minerals to eliminate measurement noise introduced by motion blur as follows:
\begin{equation}
    p_n = a \cdot p_{n-1} + (1-a) \cdot p_n^{obs}
\end{equation}
where $p_n$ and $p_{n-1}$ are estimated mineral pose at video frame $n$ and $n-1$, $p_n^{obs}$ is the raw output of the PnP solver at $n$, and $a\in[0, 1]$ is a filter coefficient.

In summary, SMMS mitigates the impact of motion blur as follows: image enhancement compensates for contrast loss, improving the recall rate; data augmentation boosts the classifier's generalization ability, and the rejection mechanism eliminates false recognitions, maintaining classification accuracy; data smoothing ensures pose estimation precision and smoothness. Through SMMS, we successfully bridge the sim-to-real gap of sensing discrepancies in the 2024 RSC. 

\section{Challenge II: Servo with Actuation Discrepancies} \label{servo system}

\begin{figure}[h]
    \centering
    \subfigure[]{
        \includegraphics[width=0.45\columnwidth]{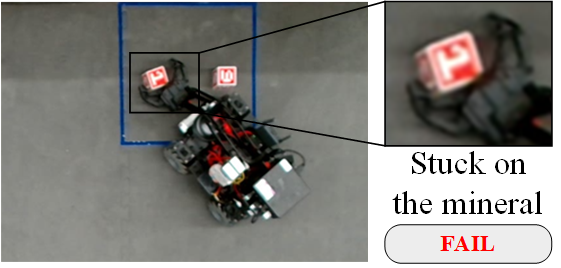}
        \label{fig:real-grasp}
    }
    \hfill
    \subfigure[]{
        \includegraphics[width=0.46\columnwidth]{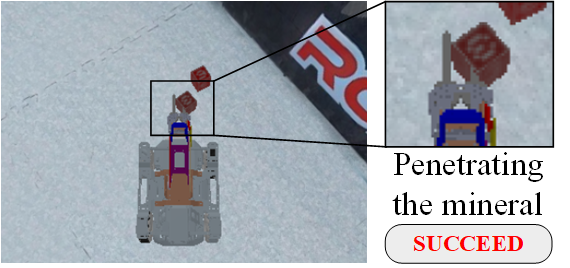}
        \label{fig:sim-grasp}
    }
    \caption{Incorrect grasp results caused by inappropriate grasp pose. When grasping the mineral diagonally in practice (a), the gripper tends to get stuck on the corner and fail. However, it "penetrates" into the mineral and succeeds in the simulator (b).}
    \label{fig:real-sim grasp}
\end{figure}

With real-time pose estimates available, the second step for the robotic system is to robustly and precisely align the gripper with the target mineral. However, there are two main actuation discrepancies in mainstream simulators. The first is inappropriate grasp poses, as illustrated in Fig. \ref{fig:real-sim grasp}, which may be the most notorious cause of inconsistent grasp results during sim-to-real transfer \cite{pyrep, ai2thor, habitat}. The second is the nonlinearity in actual dynamics, in which mainstream simulators fall short \cite{ai2thor, minos}. 
To address both, we first integrate angular alignment into chassis servo control with feedback linearization. Then we introduce Design Function (DF) to fulfill expected attenuation mode, thereby achieving necessary robustness.

\subsection{Modeling Chassis}

\begin{figure}[htbp]
\centerline{\includegraphics[width=0.9\columnwidth]{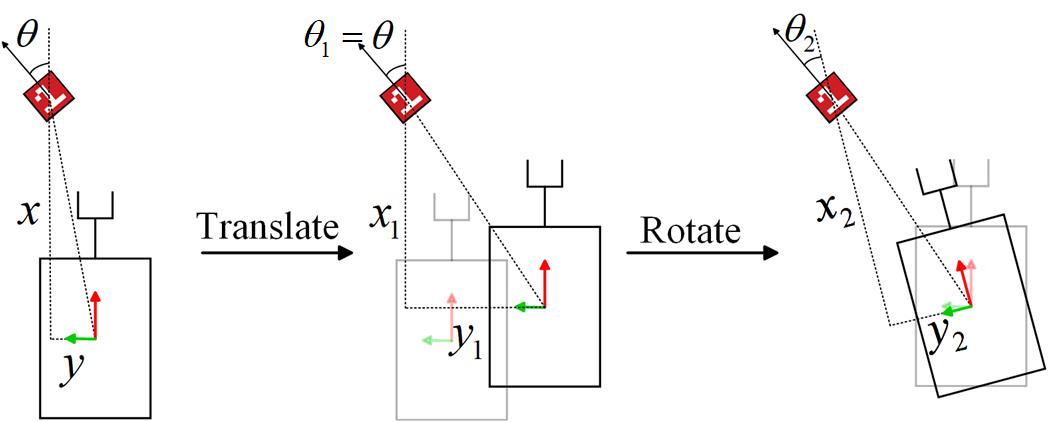}}
\caption{Marginal pose change induced by translation and rotation. The $x$ and $y$ axes of the chassis coordinate system are shown in red and green, respectively.}
\label{fig:cart-move}
\end{figure}

Let the pose of the target in the chassis coordinate system be $(x, y, \theta)^T$, and the velocity input to the omnidirectional chassis be $(v_x, v_y, \omega)^T$. As illustrated in Fig. \ref{fig:cart-move}, positional and angular changes in marginal time increment $\delta t$ induced by $v_x$, $v_y$, $\omega$ are 
\begin{equation}
    \left( \begin{matrix}
   {{x}_{2}}  \\
   {{y}_{2}}  \\
\end{matrix} \right)=\left( \begin{matrix}
   \cos \delta \theta  & \sin \delta \theta \\
   -\sin \delta \theta  & \cos \delta \theta \\
\end{matrix} \right)\left( \begin{matrix}
   x + v_x \delta t  \\
   y + v_y \delta t \\
\end{matrix} \right)
\end{equation}
and
\begin{equation}
    \theta_2=\theta - \omega \delta t
\end{equation}
Then by letting $\delta t \rightarrow 0$, we obtain the derivative of state variables as follows:
\begin{equation}
\left( \begin{aligned}
  & {\dot{x}} \\ 
 & {\dot{y}} \\ 
 & {\dot{\theta }} \\ 
\end{aligned} \right)=\left( \begin{aligned}
  & {{v}_{x}}-y\omega  \\ 
 & {{v}_{y}}+x\omega  \\ 
 & -\omega  \\ 
\end{aligned} \right)
\label{eq: state-space}
\end{equation}

\subsection{Feedback Linearization}

Here, we apply feedback linearization through inverse dynamics \cite{feedback-linearization}.
For any given trajectory $(x_d(t), y_d(t), \theta (t))^T, t\in\mathbb{R}$, the state error of the system can be denoted as $\mathbf{e}=(x-x_d, y-y_d, \theta - \theta_d)^T$ \footnote{For brevity, we abbreviate $\mathbf{e}(t)$ to $\mathbf{e}$ and use $\text{exp}\{\cdot\}$ to denote the exponential function. The notation also holds for $x$, $x_d$, etc.}. To attenuate it to 0, the derivative of the state error can be designed as follows:
\begin{equation}
\mathbf{\dot{e}}=\mathbf{f}(\mathbf{e},t)
\label{eq: DF}
\end{equation}
where $\mathbf{f}={{\left( {{f}_{1}},{{f}_{2}},{{f}_{3}} \right)}^{T}}$ is the Design Function. Substituting (\ref{eq: DF}) into (\ref{eq: state-space}), we conclude that it's fully actuated, i.e., for any expected attenuation mode determined by $\mathbf{f}$, there exists a control input
\begin{equation}
{{\left( \begin{matrix}
   {{v}_{x}} & {{v}_{y}} & \omega   \\
\end{matrix} \right)}^{T}}={{\left( \begin{matrix}
   {{f}_{1}}-y{{f}_{3}} & {{f}_{2}}+x{{f}_{3}} & -{{f}_{3}}  \\
\end{matrix} \right)}^{T}}
\end{equation}

Theoretically, $\mathbf{f}$ can be any known function of $\mathbf{e}$ and $t$. Here, we demonstrate three intuitive choices of $\mathbf{f}$: 1) multiplication of $\mathbf{e}$ (for exponential mode); 2) power function of $\mathbf{e}$ (for power-law mode); 3) linear combination of $\mathbf{e}$ and its integral (for multiple exponential modes).

\subsection{Design Function} \label{DF-comparison}

We solve the attenuation mode of state error under the above three DF choices and analyze their servo accuracy under dead zone, respectively.

For simplicity, we choose the three components of $\mathbf{f}$ as the same, where we only need to analyze in scalar form of (\ref{eq: DF}):
\begin{equation}
\dot{e}=f(e,t)
\label{eq: DF-scalar}
\end{equation}
where $e$ can represent positional or angular error. Without loss of generality, we consider position control.
In order for $e$ to have an analytical solution, we can take $f$ of the following form. (controller-specific parameters $k_p$, $k_i$ and $\alpha$ are positive real numbers)

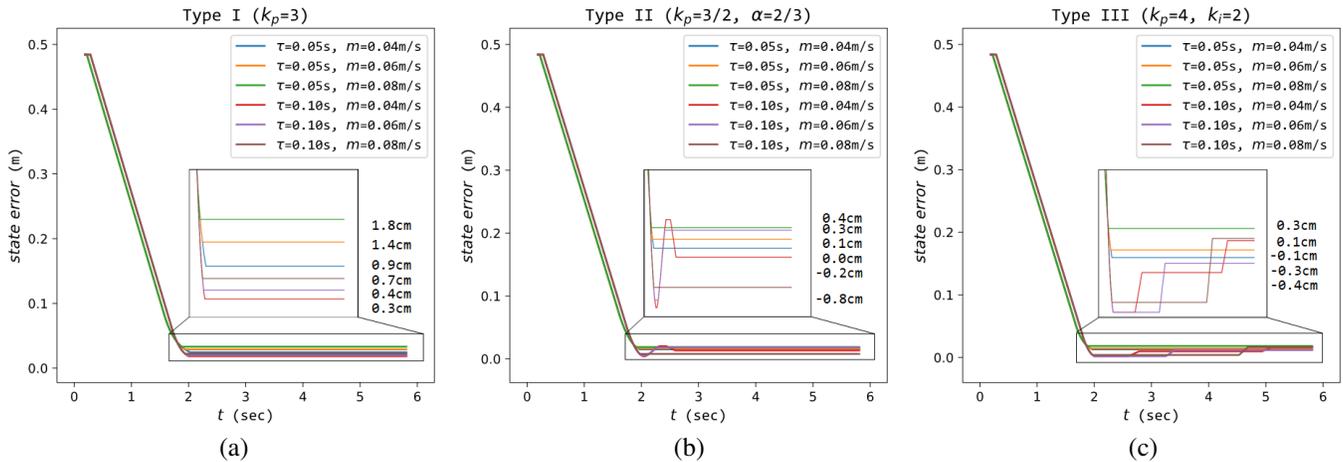
\begin{figure*}[htbp]
\centerline{\includegraphics[width=\textwidth]{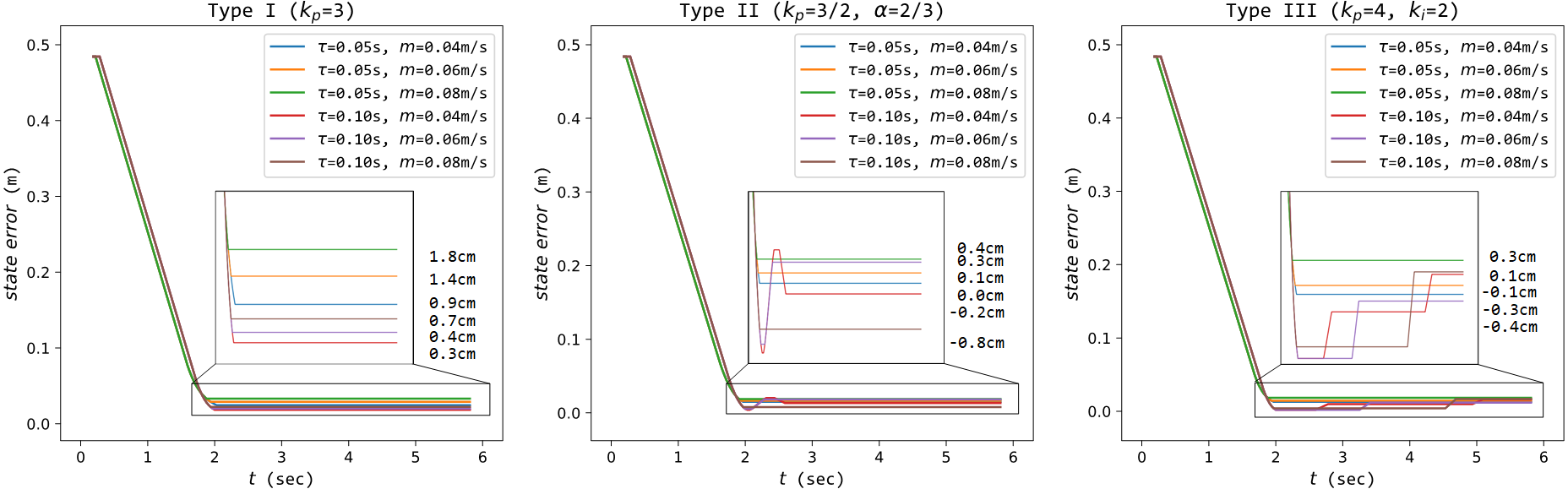}}
\begin{picture}(0,0)
    \put(0.16\linewidth,0){(a)}
    \put(0.5\linewidth,0){(b)}
    \put(0.84\linewidth,0){(c)}
\end{picture}
\caption{Error attenuation curves of the servo system with three DF choices, where the transport delay $\tau \in$[0.05, 0.1], the velocity dead zone size $m\in[0.04, 0.08]$ and the velocity saturation is 0.3 m/s. Steady-state errors are annotated on each curve. Type I controller (a) is sensitive to dead zone and fails to achieve sub-centimeter accuracy; Type II controller (b) barely meets the requirement; Type III controller (c) meets the requirement with an accuracy margin of more than 0.5 cm.}
\label{fig:controller-comparison}
\end{figure*}

\begin{enumerate}

\item[I.]{$f(e,t)=-{{k}_{p}}e$}

We can immediately solve from (\ref{eq: DF-scalar}) that $e=e(0)\cdot {{e}^{-{{k}_{p}}t}}$, indicating that the state error decays exponentially to 0.

let $m$ denote the size of velocity dead zone, i.e., 
\begin{equation}
    \Gamma (v)=\left\{ \begin{matrix}
   0 & |v|\le m  \\
   v & |v|>m  \\
\end{matrix} \right.
\end{equation}
where $v$ is the velocity, $\Gamma$ is the dead zone. Then the steady-state error ${{e}_{ss}}$ satisfies $m=\left| -{{k}_{p}}{{e}_{ss}} \right|$, from which we get  ${{e}_{ss}}=\frac{m}{{{k}_{p}}}$.

Type I controller offers a concise form and simplifies computation. However, the steady-state error is linearly dependent on the dead zone size, which renders the controller fragile.

\item[II.]{$f(e,t)=-{{k}_{p}}{{e}^{\alpha }},\,\,\,\,0<\alpha <1$} 

Assume $e(t)>0$ \footnote{In order to extend the power function to $e(t)<0$, we only need to let $f(e,t)=-{{k}_{p}}\cdot \text{sign}(e)\cdot {{\left| e \right|}^{\alpha }}$ and set $\tilde{e}=-e$, which yields symmetric results.}. By substituting this DF into (\ref{eq: DF-scalar}) and integrating after separating the variables, we have $e={{\left( {e(0)}^{1-\alpha }-(1-\alpha ){{k}_{p}}t \right)}^{\frac{1}{1-\alpha }}}$ or $e=0$.

Likewise, ${{e}_{ss}}$ satisfies $m=\left| -{{k}_{p}}{{e}_{ss}}^{\alpha } \right|$, from which we get ${{e}_{ss}}={{\left( \frac{m}{{{k}_{p}}} \right)}^{\frac{1}{\alpha }}}$.

The rationale of Type II controller is to boost the velocity output when the error is small. In practice, $m$ is around 0.05 m/s, $\alpha$ is taken as 2/3, and $k_p$ is often taken as 1, which yields a much smaller $e_{ss}$ than Type I.

\item[III.]{$f(e,t)=-{{k}_{p}}e-{{k}_{i}}\int_{0}^{t}{e(\tau )d\tau }$}

Substituting this DF into (\ref{eq: DF-scalar}) and letting $\epsilon=\int_{0}^{t}{e(\tau )d\tau }$, we obtain a second-order differential equation: 
\begin{equation}
\ddot{\epsilon}+{{k}_{p}}\dot{\epsilon}+{{k}_{i}}\epsilon=0    
\label{eq:ode}
\end{equation}
When $k_p^2-4k_i\neq0$, equation (\ref{eq:ode}) has two different roots $\lambda_{1,2}\in\mathbb{C}$. Then we substitute the initial condition $\left\{ \begin{aligned}
  & \epsilon(0)=0 \\ 
 & \dot{\epsilon}(0)=e(0) \\ 
\end{aligned} \right.$ and differentiate $\epsilon$ with respect to $t$, which yields $e=\frac{-\lambda_1}{\lambda_2-\lambda_1}e(0)\text{exp}\{\lambda_1t\}+\frac{\lambda_2}{\lambda_2-\lambda_1}e(0)\text{exp}\{\lambda_2t\}$, indicating two exponential modes of the system .

By introducing integral term, Type III controller manages to eliminate steady-state error for any dead zone size, exhibiting much stronger robustness to dead zones compared to the Type I controller.

\end{enumerate}

In summary, while the Type I controller is the simplest, it lacks the robustness required for sim-to-real transfer. The Type II controller improves upon Type I in terms of accuracy, with the requirement barely met yet. The Type III controller stands out as the most robust and accurate among the three. In further analysis with transport delay, dead zone and saturation, it provides an accuracy margin of more than 0.5 cm (see Fig. \ref{fig:controller-comparison} for extensive results).

Therefore, with feedback linearization and Design Function, we manage to build a nonlinearity-robust servo system that handles inappropriate grasp pose and unmodeled nonlinearities simultaneously, which proves to be effective in the 2024 RSC.

\section{Experiments}

In this section, we evaluate our visual perception system, servo system, and the entire robotic system within the simulator and real-world. The robot is equipped with Intel$^\circledR$ NUC11PAHi7 with 8 GB of memory and without dedicated GPU. The simulation runs in Docker containers to match the computing power of real hardware. The real-world test results are based on the final evaluation conducted by the competition committee. Original data and multi-view videos of the evaluation are available at \url{https://drive.google.com/drive/folders/1yd4S0y6_FNU45j6xjc7f12Y2G_dM1_Cl}.

\subsection{Evaluating Visual Perception System} \label{eval-perception}

\textbf{Evaluation Metrics.} Following common practices, we use recall rate, error rate and sample standard deviation as metrics. Recall rate is defined as the ratio of the number of recognized minerals (those rejected by the classifier are excluded) to the total number of minerals of frames. Error rate is the ratio of the number of misclassified contours to the number of the recognized.

Let $\mathbf{p}_i$ and $\theta_i$ be the estimated position and orientation angle (the angle between the surface normal and the positive $x$-axis) of the target mineral in the global coordinate system at time step $i$, with $\bar{\mathbf{p}}$ and $\bar{\theta}$ being their mean values. For the temporal sequence of estimates of length $l$, we use the sample standard deviation to evaluate its precision:
\begin{equation}
{{S}_{pos}}=\sqrt{\frac{1}{l-1}\sum\limits_{i=1}^{l}{{\left\| \mathbf{p}_i-\bar{\mathbf{p}} \right\|}_2^2}}
\end{equation}

\begin{equation}
{{S}_{att}}=\sqrt{\frac{1}{l-1}\sum\limits_{i=1}^{l}{{{(\theta_i-\bar{\theta})}^{2}}}}
\end{equation}

\textbf{Results.} We train our classifier for 1200 epochs with Adam optimizer and set the rejection threshold as 12 to achieve the complete rejection of incorrect contours. Then, we test recall rate, error rate and pose estimation precision within 1 m (see Table \ref{tab:perception-perf}).

\begin{table}[htbp]
\begin{center}
\begin{threeparttable}
\caption{Performance of Visual Perception Methods in 2024 RSC Simulated and Real-World Environments}
\renewcommand{\arraystretch}{1.25} 
\setlength{\tabcolsep}{8pt} 
\begin{tabular}{|c|c|c|c|c|c|}
    \hline
    \multirow{2}{*}{Method}&\multirow{2}{*}{Env}&Recall&Error&\multicolumn{2}{|c|}{Precision} \\
    \cline{5-6}
    &&Rate\tnote{*}&Rate\tnote{*}&$S_{pos}$&$S_{att}$ \\
    \hline
    Vanilla&Sim&48.7\%&1.1\%\tnote{\dag}&1.4 cm&4.8 deg \\
    \cline{2-6}
    ArUco&Real&49.2\%&1.1\%\tnote{\dag}&1.2 cm&2.1 deg \\
    \hline
    
    Ours&Sim&96.0\%&0.0\%\tnote{\dag}&1.3 cm &1.0 deg \\
    \cline{2-6}
    w/o SMMS&Real&50.5\%&0.0\%\tnote{\dag}&1.2 cm&2.2 deg \\
    \hline
    
    Ours&Sim&91.6\%&0.0\%&0.8 cm&1.0 deg \\
    \cline{2-6}
    w/ SMMS&Real&70.0\%&0.0\%&0.6 cm&1.9 deg \\
    \hline
\end{tabular}
\begin{tablenotes}
    \item[*] represents the average of the measurements for mineral markers and the exchange markers.
    \item[\dag] represents the results susceptible to detector parameters and the background. (e.g. brown doors in the room where competition holds cause about 5\% false recognitions, thereby leading to notable misclassifications)
\end{tablenotes}
\label{tab:perception-perf}
\end{threeparttable}
\end{center}
\end{table}

Compared with vanilla ArUco detector, our perception pipeline increases recall rate of minerals (from 48.7\% to 96.0\%) and reduces error rate of classification (from 1.1\% to 0.0\%), both of which provide more valid samples for pose estimation, thereby raising pose estimation precision. However, without SMMS, it suffers great loss of recall rate (from 96.0\% to 50.5\%), and falls short in dealing with false recognitions caused by visible objects in the background. By applying SMMS, our visual perception system reduces recall rate loss during sim-to-real, and enhances the robustness by rejecting all false recognitions, which yields pose estimation precision within 1 cm and 2 deg.

Additionally, we test the run-time of each component of our pipeline. It shows that the average run-times of our detector, classifier and PnP solver are 6.68, 3.60 and 0.86 ms per frame (11.14 ms per frame in total), exhibiting high efficiency of our visual perception system.

Overall, with proposed SMMS, our visual perception system manages to address measurement noise introduced by motion blur. It mitigates the sharp drop in recall rate and maintains classification accuracy and estimation precision, demonstrating noise resilience and high consistency during sim-to-real transfer.

\subsection{Evaluating Servo System}
We evaluate the accuracy and speed of our servo system using the Type III controller, which is identified as the most robust and accurate in section \ref{DF-comparison}. The proportional gain $k_p$ is set to 4, and the integral gain $k_i$ is set to 2, with an integral separation threshold\footnote{The integral term only works within given threshold.} of 0.1 m.

\textbf{Performance metrics.} We use final error and average aligning time as performance metrics. For each servo adjustment, we first define the termination condition as the position error remaining within the tolerable range for 2 seconds, with horizontal and vertical tolerances set to 0.015 m. Then, we measure aligning time (the interval from the beginning to termination, and 2 seconds of judging termination are not included) and final error (the position or attitude error when terminated, and it is marked as '-' if not aligned by the controller).\footnote{As the robot is near the target and almost stationary at the end of the alignment, measurement error introduced by the visual perception system is negligible.}

\textbf{Results.} Table \ref{tab:servo-perf} shows the average servo error and aligning time. The open-loop controller fails overcoming actual disturbance in dynamics, while the PID controller falls short in aligning coupled position and attitude simultaneously, both of which fail in real grasping. With feedback linearization and DF, our servo system with Type III controller achieves sub-centimeter accuracy with tolerable aligning time of about 4 seconds. Overall, it manages to overcome sensing discrepancies with comparable performance during sim-to-real.

\begin{table}[htbp]
\caption{Performance of Servo Control Methods in 2024 RSC Simulated and Real-World Environments}
\renewcommand{\arraystretch}{1.25} 
\setlength{\tabcolsep}{5.5pt} 
\begin{center}
\begin{threeparttable}
    \begin{tabular}{|c|c|c|c|c|c|c|}
        \hline
        \multirow{2}{*}{Method}&\multirow{2}{*}{Env}&\multicolumn{3}{|c|}{Final Error}&Aligning&\multirow{2}{*}{Result} \\
        \cline{3-5}
        &&$x$ (cm)&$y$ (cm)&$\theta$ (deg)&Time (s)& \\
        \hline
        \multirow{2}{*}{Open-loop\tnote{*}}&Sim&1.24&0.13&-&2.16&$\checkmark$ \\
        \cline{2-7}
        &Real&7.11&4.07&-&Timeout&$\times$ \\
        \hline
        \multirow{2}{*}{PID\tnote{*}}&Sim&0.27&0.22&-&2.18&$\checkmark$ \\
        \cline{2-7}
        &Real&0.35&0.86&-&Timeout&$\times$ \\
        \hline
        Ours&Sim&0.48&0.40&0.61&4.06&$\checkmark$ \\
        \cline{2-7}
        (Type III)&Real&0.81&0.63&2.18&4.12&$\checkmark$ \\
        \hline
    \end{tabular}
    \begin{tablenotes}
        \item[*] represents unable to align attitude while aligning position.
    \end{tablenotes}
\end{threeparttable}
\label{tab:servo-perf}
\end{center}
\end{table}
\subsection{Full System Evaluation in 2024 RSC}
We evaluate the grasping and placing performance of the entire robotic system in both simulation and reality. 

\textbf{Results.} Table \ref{tab:overall-perf} shows the ratio of successful grasps to total attempts during the grasping and placing processes. During the final evaluation of the 2024 RSC, our system achieved a 100\% success rate in both grasping and stacking tasks, being the only team to successfully complete both two-layer and three-layer stacking.

\textbf{Postmortem.} We observed approximately a 1/6 failure rate in placing tasks in simulation. Further analysis reveals that these failures are due to the minerals falling after being stacked. This issue arises because the simulator cannot handle collisions effectively, leading to mineral jitter and dropping. 

\begin{table}[htbp]
\caption{Pick-and-place Success Rate in 2024 RSC Simulated and Real-World Environments}
\renewcommand{\arraystretch}{1.25} 
\setlength{\tabcolsep}{18pt} 
\begin{center}
\begin{tabular}{|c|c|c|}
    \hline
    Environment&Grasp Success&Stack Success \\
    \hline
    Simulator&30/30&22/30 \\
    \hline
    Reality&6/6&6/6 \\
    \hline
\end{tabular}
\label{tab:overall-perf}
\end{center}
\end{table}

\section{Conclusion and Discussion}


\textbf{Conclusion.} In this paper, by applying Sequential Motion-blur Mitigation Strategy (SMMS), our visual perception system manages to address motion blur and provide consistently precise pose estimates; by applying feedback linearization and introducing Design Function (DF), our servo system manages to address inappropriate grasp pose as well as unmodeled nonlinearities, and maintain sub-centimeter accuracy of servo control. Additionally, we provide extensive analysis and comparison of three intuitive DF choices and conclude that the Type III demonstrates required robustness to nonlinearities. Benefiting from these, our robotic system successfully completed long-horizon pick-and-place tasks both in simulator and in real-world, securing first place in the 2024 Robotic Sim2Real Challenge \cite{RSC2024}.

\textbf{Discussion.} From the perspective of robotic systems, the presented engineering effort serves as a proof-of-concept demonstration, with its components inherently designed to meet competition-specific objectives. As such, it is not directly transferable to industrial production environments without further adaptation. However, from a sim-to-real standpoint, our robotic system effectively demonstrates the feasibility of achieving consistent performance across both simulated and real-world environments, particularly in long-horizon pick-and-place tasks.

\section*{Acknowledgment}
The authors wish to sincerely thank Associate Researcher Haoran Li and their colleagues for their constructive suggestions regarding the experimental work. We also extend our gratitude to the Deep Reinforcement Learning Group for providing the EP robot used in this research.

\end{document}